\documentclass[letterpaper, 10 pt, journal, twoside]{IEEEtran}
\usepackage{amsmath,amsfonts} 
\usepackage{algorithmicx}
\usepackage{algorithm}
\usepackage[noend]{algpseudocode}
\usepackage{array}
\usepackage[caption=false,font=normalsize,labelfont=sf,textfont=sf]{subfig}
\usepackage{textcomp}
\usepackage{stfloats}
\usepackage{url}
\usepackage{verbatim}
\usepackage{graphicx}
\usepackage{cite}
\usepackage{booktabs}
\usepackage{blindtext}

\usepackage{booktabs}  
\usepackage{multirow}  
\usepackage{amsmath}   
\usepackage{adjustbox} 

\usepackage{color,soul}

\begin{document}

\title{Model selection and real-time skill assessment for suturing in robotic surgery}



\author{Zhaoyang Jacopo Hu$^{1}$, Alex Ranne$^{2}$, Alaa Eldin Abdelaal$^{3}$, Kiran Bhattacharyya$^{4}$, \\ Etienne Burdet$^{5}$, Allison M. Okamura$^{3}$, Ferdinando Rodriguez y Baena$^{1}$

\thanks{
This work was supported by the EPSRC and Intuitive Surgical in the form of an industrial CASE studentship, the Hamlyn Centre, UKRI CDT in AI for Healthcare under
Grant EP/S023283/1, and the Stanford Institute for Human-Centered Artificial Intelligence.
\it{(Corresponding authors: Zhaoyang Jacopo Hu, Etienne Burdet, Ferdinando Rodriguez y Baena)}}
\thanks{$^{1}$Zhaoyang Jacopo Hu and Ferdinando Rodriguez y Baena are with the Department of Mechanical Engineering, Imperial College London, SW7 2AZ, (e-mail: \{jacopo.hu20, f.rodriguez\}@imperial.ac.uk).}
\thanks{$^{2}$Alex Ranne is with the Department of Computing, Imperial College London, SW7 2AZ.}
\thanks{$^{3}$Alaa Eldin Abdelaal and Allison M. Okamura are with the Department of Mechanical Engineering, Stanford University, Stanford, CA 94305 USA.}
\thanks{$^{4}$Kiran Bhattacharyya is with Intuitive Surgical, Inc., Data and Analytics, Norcross, Georgia, USA.}
\thanks{$^{5}$Etienne Burdet is with the Department of Bioengineering, Imperial College London, W12 0BZ, United Kingdom (e-mail: e.burdet@imperial.ac.uk).}
\thanks{Digital Object Identifier (DOI): see top of this page.}
}

\markboth{ } 
{Hu \MakeLowercase{\textit{et al.}}: Model selection and real-time skill assessment for suturing in robotic surgery
}


\maketitle

\begin{abstract}
Automated feedback systems have the potential to provide objective skill assessment for training and evaluation in robot-assisted surgery. In this study, we examine methods to achieve real-time prediction of surgical skill level in real-time based on Objective Structured Assessment of Technical Skills (OSATS) scores. Using data acquired from the da Vinci Surgical System, we carry out three main analyses, focusing on model design, their real-time performance, and their skill-level-based cross-validation training. 
For the model design, we evaluate the effectiveness of multimodal deep learning models for predicting surgical skill levels using synchronized kinematic and vision data. 
Our models include separate unimodal baselines and fusion architectures that integrate features from both modalities and are evaluated using mean Spearman's correlation coefficients, demonstrating that the fusion model consistently outperforms unimodal models for real-time predictions.
For the real-time performance, we observe the prediction's trend over time and highlight correlation with the surgeon's gestures.
For the skill-level-based cross-validation, we separately trained models on surgeons with different skill levels, which showed that high-skill demonstrations allow for better performance than those trained on low-skilled ones and generalize well to similarly skilled participants. 
Our findings show that multimodal learning allows more stable fine-grained evaluation of surgical performance and highlights the value of expert-level training data for model generalization.

\end{abstract}


\section{Introduction}
\IEEEPARstart{S}{kill} assessment in robotic surgery is a necessary component to provide feedback in surgical training and ensuring patient safety during procedures. With the increasing adoption of robotic systems like the da Vinci Surgical System, there is a growing demand for reliable, objective, and efficient methods to evaluate surgical performance. Traditional approaches to skill assessment \cite{chen2019objective} rely on post-procedure analysis, often involving subjective expert evaluations and manual review of recorded data. While these methods provide valuable insights, they lack immediacy and scalability, making it challenging to offer real-time feedback to surgeons during procedures.

\begin{figure}[t] 
\centering 
\includegraphics[height=3.3cm, trim={0.0cm 0.0cm 0.0cm 0.0cm},clip]{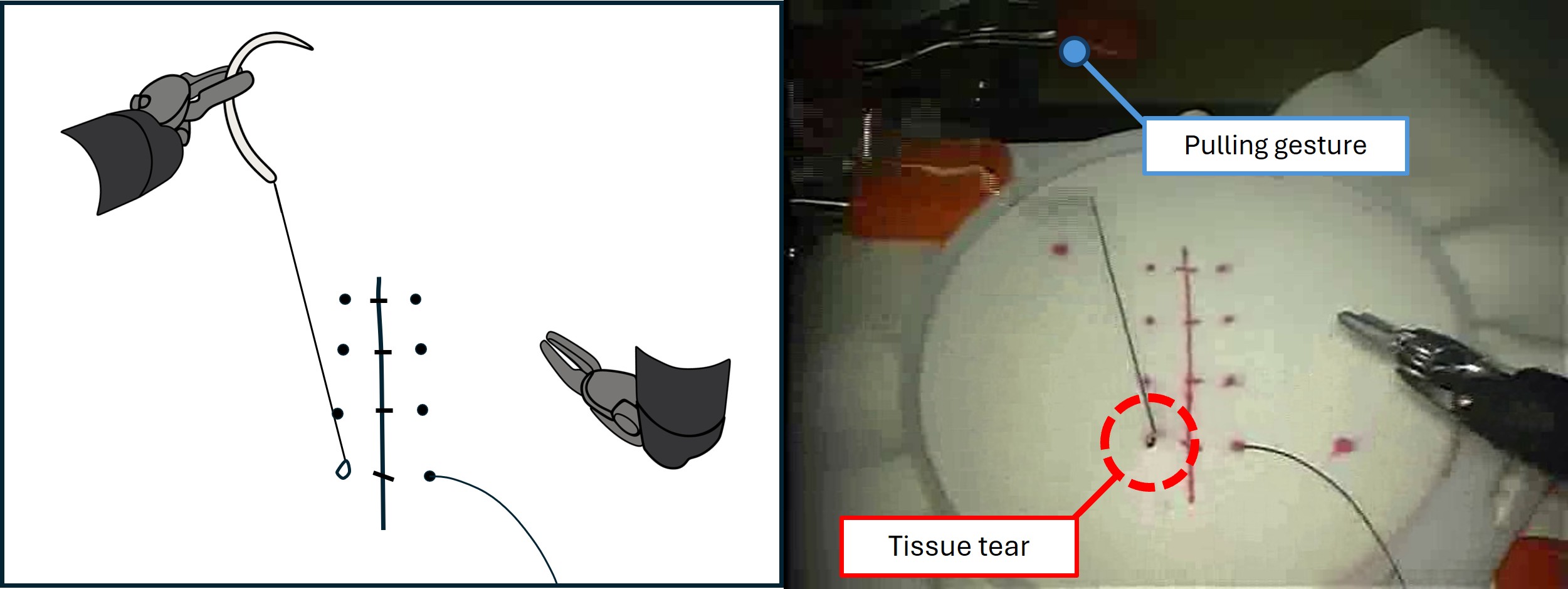}
\caption{Suturing example in the JIGSAWS dataset. As the needle is pulled out, the tissue is visibly torn, which affects the skill assessment score ``Respect for Tissue".} 
\label{fig: pulling sample}
\end{figure} 

To address this limitation, researchers have developed machine learning models capable of automatically assessing surgical skill. A widely used resource for training such models is the JHU-ISI Gesture and Skill Assessment Working Set (JIGSAWS) \cite{gao2014jhu}. JIGSAWS includes multimodal data captured during robotic suturing tasks (Figure \ref{fig: pulling sample}), combining synchronized vision and kinematic data streams. The dataset is widely used as a common standard for investigating surgeon movements during robotic surgery \cite{van2021gesture, van2022gesture, van2020multi}.
However, in the context of skill assessment, the annotations of technical skills in these types of datasets are limited, as they typically use only a single score per trial.
This constraint has led most existing studies to focus on post-hoc evaluations, providing only trial-level assessments that overlook the specific gestures or events correlated to high or low skills \cite{zia2018automated, benmansour2023deep, guo2024mt}.

To address this, our work introduces real-time, window-level skill prediction models that offer fine-grained, time-resolved feedback throughout the surgical task, enabling a shift toward continuous and adaptive evaluation.
From the clinical perspective, having real-time skill assessment allows trainees to adjust their movements during a procedure, which is more effective in long term training \cite{laca2022using}.
A second limitation concerns the diversity and imbalance of participant expertise within such datasets. Recruiting expert surgeons is difficult, often resulting in datasets that contain a broad range of skill levels. This raises concerns about the reliability of labels and the influence of user skill level on model performance \cite{hashemi2023acquisition}. To investigate this, we conduct skill-level-based cross-validation, examining how models trained on participants of varying expertise perform and generalize across different user groups.
Lastly, existing research often treats vision and kinematic data in isolation \cite{benmansour2023deep, menegozzo2019surgical, funke2019video, ding2023sedskill}, developing separate models for each modality and overlooking their complementary nature. To overcome this, we explore both unimodal and multimodal model architectures by evaluating seven deep learning models that process kinematic, visual, and combined input streams.

The contributions of this work are as follows: 
(1) We present a comparative analysis of seven models across unimodal (kinematic or vision) and multimodal (fusion) approaches, establishing a foundation for understanding the effectiveness of different data modalities;
(2) We extend these models to perform real-time, granular skill assessment, shifting away from post-hoc evaluations and aligning more closely with how surgical performance is assessed in live procedures;
(3) We examine how varying levels of surgical expertise affect model performance, providing insight into the generalizability and robustness of skill assessment systems across different user skill levels.

Together, these contributions form a framework for evaluating and enhancing real-time, multimodal surgical skill assessment, addressing both model effectiveness and practical deployment considerations.




\section{Related Work}

Surgical skill assessment has commonly relied on an expert reviewing videos of the surgical task and manually scoring the performance of the surgeons. Zia et al. \cite{zia2018automated} highlighted the biased and time-consuming nature of manual assessments, emphasizing the need for automation.

Several approaches have been explored to automate surgical skill assessment.
Early methods leveraged Hidden Markov Models (HMMs) for modeling surgical gestures and classifying skill levels \cite{tao2012sparse}. More recent studies have applied deep learning architectures, such as convolutional and recurrent neural networks (CNNs, RNNs), to extract spatiotemporal features from kinematic and video data. For instance, Zia et al. \cite{zia2018automated} introduced a feature fusion approach using sequential motion texture, discrete Fourier transform, and discrete cosine transform to classify skill levels in the JIGSAWS dataset. 

Several studies have also focused on categorical skill classification, grouping surgeons into novice, intermediate, or expert levels \cite{gao2014jhu, wang2018deep}. However, these methods often lack the granularity needed for real-time, continuous assessment.
One of the most widely adopted skill assessment frameworks is the Objective Structured Assessment of Technical Skills (OSATS), which evaluates surgical performance based on predefined criteria such as respect for tissue, time and motion, and overall performance which allows a granular study of the surgical skills. Benmansour et al. \cite{benmansour2023deep} proposed a Convolutional Neural Network and Bidirectional Long Short-Term Memory (CNN+BiLSTM) model for predicting OSATS scores directly from kinematic data, achieving state-of-the-art performance. However, the use of BiLSTM limits the architecture to only perform post-hoc analysis on the recorded data.
Additionally, many surgical skill datasets provide only a single label for an entire trial, typically assigned after task completion. While useful for global assessment, this labeling approach limits the ability to evaluate skill progression throughout the procedure. However, it is possible to leverage these trial-level labels to supervise in a finer-grained manner, enabling models to infer skill continuously over time despite the coarse annotation \cite{hira2022video}.

This work extends prior research in surgical skill assessment by introducing three primary contributions:
\begin{itemize}
\item \textbf{Performance analysis with different input modalities}: We analyze the performances of models based on different data channels using Spearman's correlation coefficient and Root Mean Square Error ($RSME$).

\item \textbf{Real-time OSATS Score Prediction}: Unlike studies that classify user skills into discrete categories (novice, intermediate, expert), our approach focuses on predicting continuous OSATS scores in real-time. This allows for a more nuanced evaluation of surgical performance throughout a procedure.

\item \textbf{Comprehensive Dataset Analysis}: Existing studies often apply machine learning models without extensive dataset analysis. We conduct a detailed study of the dataset, addressing potential limitations in the OSATS labels.
\end{itemize}
Our approach aims to enhance the precision, interpretability, and practical applicability of real-time surgical skill assessment in robotic surgery training.

\section{Data Pre-Processing}

The JIGSAWS dataset is a collection of labeled dry lab tasks performed by 8 surgeons on the da Vinci Surgical System. In this paper, we focus on the dataset's suturing data, as it is a widely studied and analyzed task for state-of-the-art models \cite{hu2025confidence, hari2024stitch}.
The dataset consists of three parts: kinematic data, video data, and manual annotations. Both the kinematic and video data are captured at 30 Hz and synchronized.
The kinematic data are collected from the da Vinci Surgical System while the video data captures the surgical scene from an endoscopic camera.
The dataset also includes skill assessments using the OSATS metrics and manual annotations for the gesture from G1 to G15, each representing the surgical activity performed by the surgeon in a segment of time. Notice that not all surgical gestures might be used in the suturing task.

\subsection{Kinematic data}

The dataset provides 76 kinematic variables (3 for Cartesian position, 9 for the rotation matrix, 3 for linear velocities, 3 for angular velocities, and 1 for gripper angle, for each manipulator) obtained from the manipulators of the da Vinci platform: the patient-side manipulators (PSM1 and PSM2) and surgeon-side tool manipulators (MTM1 and MTM2). These variables can then be used to construct the features.
To represent the interaction between two hands, Rakita et al. \cite{rakita2019shared} showed that it can be achieved using six features which can be computed as scalar-valued features to represent a single time point:

\begin{equation}\label{six scalars}
\mathbf{s}_t =
\begin{bmatrix}
\|\mathbf{p}_t^d - \mathbf{p}_t^n\| \\
\|\mathbf{p}_t^d - \mathbf{p}_t^n\| - \|\mathbf{p}_{t-1}^d - \mathbf{p}_{t-1}^n\| \\
\|\mathbf{p}_t^d - \mathbf{p}_{t-1}^d\| \\
\|\text{displacement}(\mathbf{q}_t^d, \mathbf{q}_{t-1}^d)\| \\
\|\mathbf{p}_t^n - \mathbf{p}_{t-1}^n\| \\
\|\text{displacement}(\mathbf{q}_t^n, \mathbf{q}_{t-1}^n)\|
\end{bmatrix}
\end{equation}
where $\mathbf{p}_t^d$, $\mathbf{q}_t^d$, $\mathbf{p}_t^n$, and $\mathbf{q}_t^n$ are the positions and orientations of the dominant and nondominant hands at time step $t$, respectively.
The six scalar values calculated, as shown in Equation \ref{six scalars}, produce relative coordinates (in contrast to absolute coordinate frames):
(i) hand offset (distance between hands), (ii) hand offset velocity (rate of change of hand offset), (iii) translational and (iv) rotational velocities of the dominant hand, and (v) translational and (vi) rotational velocities of the nondominant hand.
To account for variations in the kinematic features, min-max normalization was then applied to scale the features.

As in Rakita et al. \cite{rakita2019shared}, the features $\mathbf{s}_t$ only encode a single, discrete hand pose event. For the real-time implementation, we set up a window $\omega$ with many such events together in a long concatenated vector $\mathbf{f}_t$ to encode motion over time:

\begin{equation}
\mathbf{f}_t =
\begin{bmatrix}
\mathbf{s}_{t - \omega} \\
\mathbf{s}_{t - \omega + 1} \\
\vdots \\
\mathbf{s}_t \\
\end{bmatrix}
\end{equation}
Each kinematic sequence and corresponding video frame sequence were segmented using a sliding window approach. A fixed window size $\omega$ of 50 time steps with a stride of 10 was selected to maintain a fine-grained temporal resolution while ensuring overlapping contextual information. This approach enabled the model to capture temporal dependencies within both kinematic and visual data streams.

\subsection{Video data}

The video data was preprocessed to ensure consistency and facilitate model training. Each video was first converted into a sequence of frames extracted at a fixed 30 Hz frame rate to standardize temporal resolution across all samples.

Each frame was resized to 224×224 pixels to maintain consistency across samples. The images were converted into tensors and normalized using standard ImageNet mean ([0.485, 0.456, 0.406]) and standard deviation ([0.229, 0.224, 0.225]), ensuring compatibility with pretrained models. 
Finally, as mentioned, the frames were organized into fixed-length sequences, with overlapping sliding windows used to preserve temporal continuity and provide context for the model.

\subsection{OSATS rating for skill evaluation}

In the JIGSAWS's suturing dataset, the surgical skills of the participants are annotated in a standard way by using a modified OSATS approach. These are six scores manually assigned by a cardiac surgeon for each trial (``Respect for tissue", ``Suture/needle handling", ``Time and motion", ``Flow of operation
", ``Overall performance", and ``Quality of final product"), and each rating is scored on a Likert scale of 1 (Worst) through 5 (Best). 
In this paper, we use the notation OSATS$^n$ to denote the $n$-th score, and OSATS$^{n\text{-}m}$ to indicate a range of scores from $n$ to $m$, inclusive.

It is important to note that these scores are designed for post-hoc evaluation, which the literature was limited to due to the single score label per trial. As our aim is to implement a granular prediction that can be used in a real-time scoring framework, their interpretation must be adjusted to reflect the temporal nature of the evaluation. For instance, while ``Respect for Tissue" can naturally be assessed in real-time, ``Flow of Operation" should require analysis over a longer time window to provide meaningful feedback. As such, in the context of real-time scoring, these scores will be interpreted with respect to the specific time window under consideration.

\section{Models and Cross-validation}\label{Sec: model}

In this work, we implement seven machine learning models to investigate the usefulness of kinematic and video data for surgical skill assessment. Holistically, these models fall into two well-established architectural families: LSTM-based and Transformer-based networks. Each architecture is adapted to handle unimodal inputs (either kinematic or vision data) as well as fused multimodal inputs.

LSTM-based models process sequences frame by frame, capturing temporal dynamics through recurrent connections. We implement three variants:
\begin{itemize}
    \item \textit{CNN-LSTM-V} for vision-only input, where visual features maps are first extracted using a ResNet-18 CNN and then passed through LSTM layers.
    \item \textit{LSTM-K} for kinematic-only input, which directly models temporal patterns in the 18-channel kinematic sequences.
    \item \textit{DualLSTM-F} for multimodal fusion, where separate LSTMs process each modality and their outputs are fused for final prediction.
\end{itemize}

Transformer-based models process entire sequences in parallel using self-attention mechanisms to capture long-range dependencies. We implement three variants:

\begin{itemize}
    \item \textit {Transformer-V} and \textit {Transformer-K} for vision-only and kinematic-only inputs, respectively. These include self-attention across the current input feature arrays (where for Transformer-V, an additional ResNet-18 is included before the encoder), and cross-attention with previous feature arrays in time.
    \item \textit {DualTransformer-F}, a fusion model with cross-attention between kinematic and vision branches, enabling richer multimodal interaction.
\end{itemize}

Finally, we also implement a traditional vision-only baseline, \textit{CNN-V}.
For all architectures, we adopt window-based processing for real-time applicability and evaluate both unimodal and multimodal settings.

In JIGSAWS and derived studies, the two cross-validation techniques used to evaluate the performance of models are:

\begin{itemize}
    \item Leave-one-supertrial-out (LOSO): the model is trained on data from all subjects, but one trial from each subject is used for validation. The model's performance is evaluated across all participants, ensuring the robustness of the method.
    
    \item Leave-one-user-out (LOUO):  the model is trained on data from all subjects except one, which is used for testing. This process is repeated for each subject, ensuring that the model's performance is evaluated with a subject that has not previously been seen in the training dataset.
\end{itemize}

We also propose a third cross-validation technique:

\begin{itemize}
    \item Leave-one-score-in (LOSI): the model is trained on data that includes at least one matching score among the first three OSATS scores (i.e. $\exists i \in \text{OSATS}^{1\text{–}3} \text{ s.t. } i = n$), namely ``Respect for Tissue", ``Suture/Needle Handling", and ``Time and Motion". 
    For instance, if a trial has OSATS$^{1\text{-}3}$ equal to (1, 1, 2), it would be included in the training set for folds corresponding to scores 1 and 2 but excluded from folds for scores 3, 4, and 5. 
    This approach ensures that each fold contains trials with at least one overlapping score, allowing the model to generalize across similar skill levels.
\end{itemize}

In this work, we performed cross-validation in the LOSO, LOUO and LOSI manner, by dividing the data into windowed folds, leaving one fold out, then evaluating on the rest, and repeating for the next fold. Data augmentation was not implemented on the video data to avoid distorting or removing key information from within the frame, such as the suturing needle, the target suturing point, etc.

\section{Results and Discussion}

Using the neural networks architectures described in Section 
\ref{Sec: model}, we trained models for the robotic suturing task.
A total of three separate analyses were conducted to understand how high performing skill assessment models can be achieved:
\begin{itemize}
    \item \textit{Model performance analysis through $\rho$ and $RMSE$}: Using all seven models described in Section \ref{Sec: model}, the average of the mean Spearman's correlation coefficients and Root Mean Square Error of all OSATS in each trial is calculated to evaluate models with different input modalities and compare with the results in the literature.
   \item \textit{Visual highlights of real-time OSATS predictions and correlation with surgical gestures}: We used the single OSATS score per trial provided by JIGSAWS to perform granular predictions for the entire trial and observed the correlation with surgical gestures. 
   We performed real-time prediction of the single OSATS score as it is often overlooked in favor of post-hoc evaluation. By highlighting potential patterns learned across trials, we aim to extract additional information from the single label restriction and explore its use for real-time feedback.

   Due to computational constraints, we focus on representative kinematic-, vision-, and fusion-based models (i.e. LSTM-K, CNN-LSTM-V, and DualLSTM-F). 
   
    \item \textit{Analysis of LOSI cross-validation}: We performed validation with the LOSI setup using vision-only and kinematic-only models to observe learning performance across data with different user skills. 
    This analysis aims at understanding the quality of datasets that contain different skilled users, highlighting the most useful data.

\end{itemize}

The experimental evaluation was conducted by calculating the Root Mean Squared Error ($RMSE$) values between the ground truth and predicted OSATS scores and using the Spearman's correlation coefficient $\rho$, which is an evaluation metric that also provides a check for statistical significance using the $p$-value. The value $\rho$ can be within -1 and +1, with a value closer to +1 indicating a stronger correlation between the predicted and ground truth OSATS.

\subsection{Mean $\rho$ and Mean $RMSE$}

By using the cross-validation methods LOSO and LOUO proposed by \cite{gao2014jhu}, we compared the performances of the models.
Table \ref{mean SCC and RMSE} shows the results from averaging the mean Spearman's correlation coefficients and Root Mean Square Error ($\overline{\rho}$ \textbar\ $\overline{RSME}$) of all OSATS in each trial for all the suturing dataset.
Overall, it can be observed that using video data allows the models to perform better in both the LOSO and LOUO setups.
In particular, in LOSO, models that use video data (i.e. CNN-LSTM-V, DualLSTM-F, CNN-V, Transformer-V, and DualTransformer-F) can achieve a $\rho>0.8$ while keeping $RSME\le0.6$. LOUO displays a similar trend, with all models trained on video data performing $\rho>0.6$ and $RSME\le0.8$. The best performing model in LOSO is CNN-V, while in LOUO is CNN-LSTM-V. Noticeably, CNN-LSTM-V is the second best performing model in LOSO, while CNN-V is sixth.
The worst performing models are those reliant solely on kinematic data (i.e. LSTM-K and Transformer-K), with the worst performing in both LOSO and LOUO being Transformer-K. Interestingly, while the non-Transformer fusion model (i.e. DualLSTM-F) did not show improvements with respect to the model trained on video data only, the DualTransformer-F improved its performance with respect to the transformer models trained on video or kinematic data only.

The results obtained can also be directly compared with those in the literature.
Compared to the study from Zia et al. \cite{zia2018automated}, deep learning models outperformed the use of holistic features in automated skill assessment, regardless of the cross-validation scheme used.
In the LOSO setup, Benmansour et al \cite{benmansour2023deep}. showed that their kinematic-only based state-of-the-art model obtained an overall $\rho$ of 0.65 using CNN-BiLSTM. Our results on fusion models then indicate that the adoption of both video and kinematic data can further improve this performance.
Guo et al. \cite{guo2024mt}, which used a vision-only transformer model, also shows comparable results for LOSO and LOUO, with $\rho$ of 0.87 and 0.75, respectively. Our results suggest that a fusion with kinematic data could further improve the performance, especially considering the lower computational demands of kinematic-based models.

\begin{table}[h!]
\centering

\caption{Performance comparison for Suturing based on $\overline{\rho}$ \textbar\ $\overline{RSME}$, using different architectures and source data under LOSO and LOUO evaluation methods (highest performances in bold).}
\resizebox{\columnwidth}{!}{  
\begin{tabular}{lcc}
\toprule
\multirow{2}{*}{} & \multicolumn{2}{c}{\textbf{Suturing}} \\
\cmidrule(lr){2-3}
\textbf{} & \textbf{LOSO} & \textbf{LOUO} \\
\midrule
CNN-LSTM-V                & 0.861 \textbar\ 0.516 & \textbf{0.843} \textbar\ \textbf{0.553} \\
LSTM-K               & 0.642 \textbar\ 0.887 & 0.612 \textbar\ 0.881 \\
DualLSTM-F & 0.857 \textbar\ 0.522 & 0.671 \textbar\ 0.698 \\
CNN-V                & 
\textbf{0.898} \textbar\ \textbf{0.431} & 0.568 \textbar\ 0.810 \\
Transformer-V                & 0.821 \textbar\ 0.592 & 0.635 \textbar\ 0.825 \\
Transformer-K              & 0.560 \textbar\ 0.850 & -0.0133 \textbar\ 1.004 \\
DualTransformer-F              & 0.808 \textbar\ 0.606 & 0.654 \textbar\ 0.758 \\
\bottomrule
\end{tabular}
}
\label{mean SCC and RMSE}
\end{table}

In order to analyze each OSATS more accurately, Table \ref{real time SCC and RMSE} presents the model-wise comparison of performance across six OSATS criteria in LOSO and LOUO using the Spearman's correlation coefficient and Root Mean Square Error ($\rho$ \textbar\ $RMSE$) checking for statistical significance.

In the LOSO setup, CNN-V, which relies solely on visual input, achieved the highest overall performance, with $\rho$ values exceeding 0.89 in all metrics and a remarkably low RMSE.
This suggests that visual cues alone carry rich information regarding skill assessment in suturing.
The CNN-LSTM-V model, which incorporates temporal modeling of visual features, also performed well, particularly in \textit{Time and Motion} (0.894 \textbar\ 0.504) and \textit{Suture Handling} (0.885 \textbar\ 0.539), showing that temporal modeling offers benefits over static visual features.
Interestingly, the fusion model (DualLSTM-F) leveraging both visual and kinematic data demonstrated competitive and consistent performance across all metrics, indicating the advantage of multimodal inputs. All improvements were statistically significant ($p<0.05$).
In contrast, the LSTM-K model using only kinematic data showed comparatively lower correlation values across all metrics, suggesting that visual data plays a more pivotal role under LOSO.
Transformer-based models showed mixed results. The Transformer-V model showed strong performance, e.g. (0.911 \textbar\ 0.418) in \textit{Quality of Final Product}, though slightly behind CNN-V and fusion models. However, the Transformer-K model again underperformed, consistent with other kinematic-only approaches.

The LOUO setup, which is generally more challenging due to the higher variance introduced by unseen users, resulted in lower overall performance across models. Nonetheless, several models remained robust.
The CNN-LSTM-V model achieved strong performance in most metrics, notably in \textit{Overall Performance} (0.925 \textbar\ 0.415) and \textit{Suture Handling} (0.888 \textbar\ 0.536), confirming its generalization capability.
The fusion model (DualLSTM-F) retained moderate performance across metrics but did not outperform the vision-based CNN-LSTM-V. 
The DualTransformer-F model under LOUO performed relatively well in \textit{Suture Handling} (0.848 \textbar\ 0.5642) but struggled in \textit{Quality of Final Product} (0.302 \textbar\ 1.003), suggesting limitations in generalizing fusion learning across users.
The Transformer-K model failed to generalize under LOUO, with negative or near-zero correlation across most metrics (e.g., (-0.2028 \textbar\ 0.9827) in \textit{Quality of Final Product}), reinforcing the trend observed in the LOSO setup.

Overall, models utilizing visual data, especially CNN-based architectures, demonstrated superior performance across both LOSO and LOUO. Fusion models yielded competitive results, particularly under LOSO, indicating potential for a multimodal approach. Kinematic-only models consistently underperformed, and transformer-based models showed promise but require further tuning and data to achieve better performance.

\begin{table*}[th!]

\caption{Performance comparison for Suturing based on $\rho$ \textbar\ $RSME$, using different architectures and source data under LOSO and LOUO evaluation methods (highest OSATS performances in bold).*$p<0.05$}
LOSO
\centering
\begin{adjustbox}{width=\textwidth}  
\begin{tabular}{lcccccc}
\toprule
 & Respect for tissue & Suture handling & Time and motion & Flow of operation & Overall performance & Quality of final product \\
\midrule
CNN-LSTM-V & 0.8118 \textbar\ 0.515 & 0.885 \textbar\ 0.539 & 0.894 \textbar\ 0.504 & 0.870 \textbar\ 0.494 & \textbf{0.872} \textbar\ \textbf{0.508} & 0.836 \textbar\ 0.534\\
LSTM-K & 0.594* \textbar\ 0.837 & 0.612* \textbar\ 0.970 & 0.674* \textbar\ 0.971 & 0.699* \textbar\ 0.820 & 0.678* \textbar\ 0.890 & 0.597* \textbar\ 0.821\\
DualLSTM-F & 0.852* \textbar\ 0.496 & 0.831* \textbar\ 0.612 & 0.854* \textbar\ 0.538 & 0.863* \textbar\ 0.483 & 0.845* \textbar\ 0.575 & 0.894* \textbar\ 0.403 \\
CNN-V & \textbf{0.893}* \textbar\ \textbf{0.392} & \textbf{0.897}* \textbar\ \textbf{0.425} & \textbf{0.899}* \textbar\ \textbf{0.460} & \textbf{0.900}* \textbar\ \textbf{0.423} & 0.867* \textbar\ \textbf{0.508} & \textbf{0.929}* \textbar\ \textbf{0.363}\\
Transformer-V & 0.766* \textbar\ 0.583 & 0.825* \textbar\ 0.573 & 0.809* \textbar\ 0.699 & 0.791* \textbar\ 0.621 & 0.822* \textbar\ 0.624 & 0.911* \textbar\ 0.418\\
Transformer-K & 0.558* \textbar\ 0.848 & 0.638* \textbar\ 0.893 & 0.559* \textbar\ 0.902 & 0.481* \textbar\ 0.811 & 0.521* \textbar\ 0.849 & 0.605* \textbar\ 0.791\\
DualTransformer-F & 0.773 \textbar\ 0.590 & 0.865* \textbar\ 0.540 & 0.770* \textbar\ 0.733 & 0.819* \textbar\ 0.574 & 0.785* \textbar\ 0.639 & 0.835* \textbar\ 0.539\\
\bottomrule
\end{tabular}
\end{adjustbox}

LOUO
\centering

\begin{adjustbox}{width=\textwidth}  
\begin{tabular}{lcccccc}
\toprule
 & Respect for tissue & Suture handling & Time and motion & Flow of operation & Overall performance & Quality of final product \\
\midrule
CNN-LSTM-V & 0.729* \textbar\ 0.667 & \textbf{0.888}* \textbar\ \textbf{0.536} & \textbf{0.848}* \textbar\ \textbf{0.573} & \textbf{0.861}* \textbar\ \textbf{0.534} & \textbf{0.925}* \textbar\ \textbf{0.415} & \textbf{0.806}* \textbar\ \textbf{0.565} \\
LSTM-K & 0.520* \textbar\ 0.881 & 0.645* \textbar\ 0.976 & 0.513* \textbar\ 0.964 & 0.638* \textbar\ 0.801 & 0.688* \textbar\ 0.895 & 0.660* \textbar\ 0.744 \\
DualLSTM-F & \textbf{0.814}* \textbar\ \textbf{0.578} & 0.694* \textbar\ 0.741 & 0.619* \textbar\ 0.766 & 0.572* \textbar\ 0.718 & 0.700* \textbar\ 0.633 & 0.629* \textbar\ 0.735 \\
CNN-V & 0.713* \textbar\ 0.703 & 0.624* \textbar\ 0.854 & 0.500* \textbar\ 0.929 & 0.488* \textbar\ 0.810 & 0.668* \textbar\ 0.700 & 0.417* \textbar\ 0.837 \\
Transformer-V 
& 0.647* \textbar\ 0.712 & 0.705* \textbar\ 0.892 & 0.695* \textbar\ 0.730 & 0.617* \textbar\ 0.761 & 0.657* \textbar\ 0.730 & 0.492* \textbar\ 1.065\\
Transformer-K & -0.167 \textbar\ 0.970 & 0.192 \textbar\ 1.074 & 0.101 \textbar\ 1.057 & -0.034 \textbar\ 0.940 & 0.031 \textbar\ 0.996 & -0.203 \textbar\ 0.983\\
DualTransformer-F & 0.731* \textbar\ 0.727 & 0.848* \textbar\ 0.564 & 0.679* \textbar\ 0.750 & 0.638* \textbar\ 0.757 & 0.725* \textbar\ 0.680 & 0.302 \textbar\ 1.003\\
\bottomrule
\end{tabular}
\end{adjustbox}

\label{real time SCC and RMSE}
\end{table*}

\subsection{Real-time OSATS predictions}

\begin{figure*}[t] 
\centering 
\includegraphics[width=\textwidth, trim={1cm 3.3cm 1cm 3.8cm},clip]{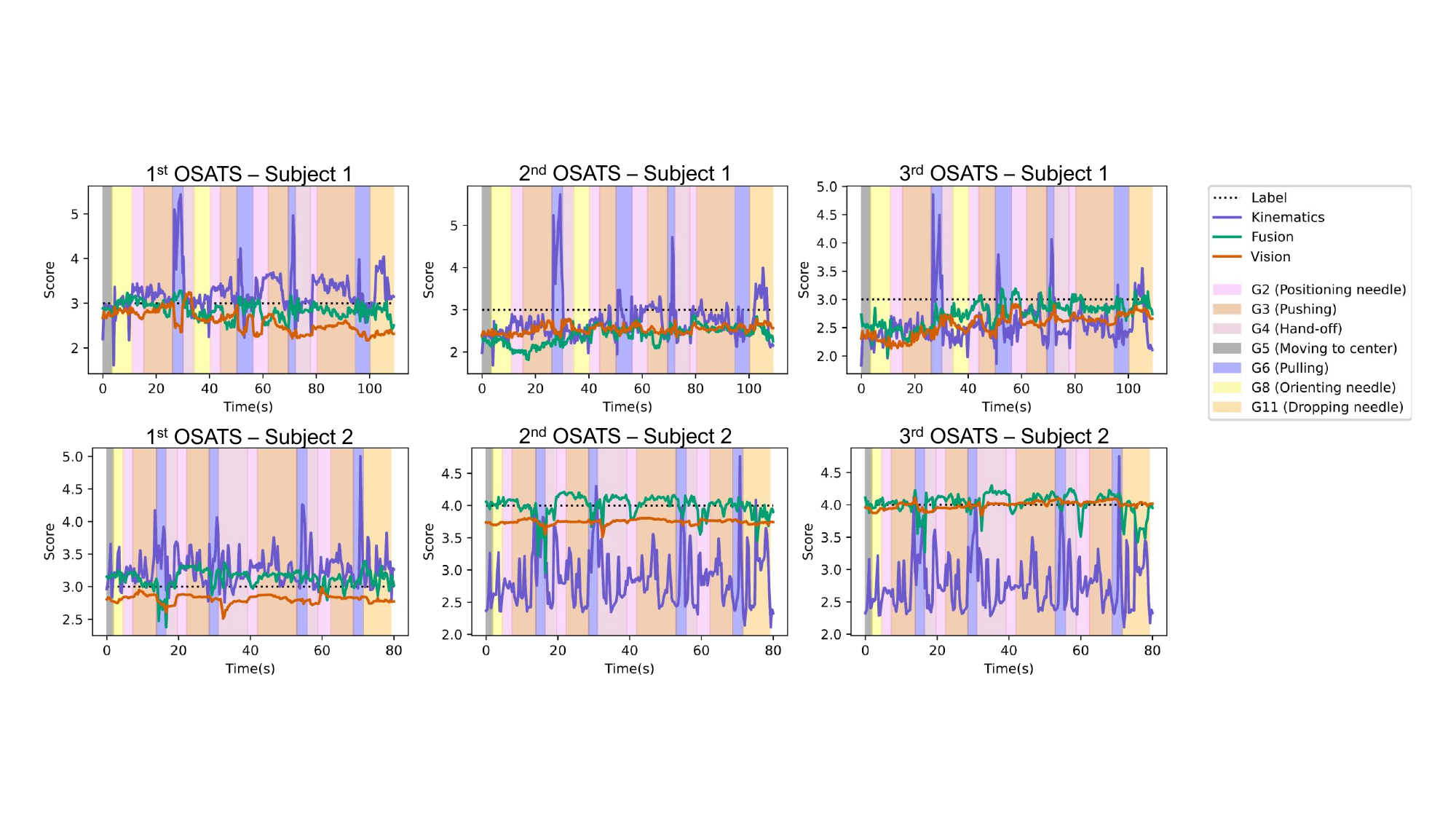}
\caption{Real-time prediction of OSATS scores for two representative sample subjects. Each subplot corresponds to a different OSATS$^{1\text{-}3}$ component, where the x-axis denotes time and the y-axis reflects the prediction scores. 
The lines represent the ground truth label, and the predictions from the kinematic-only-based model, vision-based-only model, and the fusion model. 
The background is segmented with colored bands indicating the active gesture label $G$ as provided by JIGSAWS, providing contextual cues from the temporal evolution of surgical gestures.} 
\label{fig: realtime osats}
\end{figure*}

To further illustrate the typical temporal behavior of the models, we visualize the real-time prediction trends with two representative trials in Figure \ref{fig: realtime osats}. 
Across both trials, the fusion model demonstrates consistently more stable and accurate predictions compared to unimodal baselines. In the first trial, the fusion outputs closely follow the ground truth labels, even in gesture-rich segments involving frequent transitions between G2, G3, and G6. While the kinematic model exhibits high variance and the vision model tends to underestimate, the fusion model smooths over these variations, suggesting it effectively leverages complementary cues from both modalities.

The benefits of fusion become even more evident in the second trial. Here, the kinematic model again produces noisy predictions, particularly during prolonged pulling (G6) and pushing (G3) phases, where the amplitude fluctuates. The vision model, while smoother, remains insensitive to certain transitions, leading to flattened predictions across gestures. In contrast, the fusion model maintains a higher alignment with the ground truth and adapts more sharply to gesture changes. 

These qualitative results support our quantitative findings and highlight the robustness of the fusion model in dynamic surgical scenarios. The model's ability to maintain label alignment and smooth fluctuations in real time suggests strong potential for future applications in surgical skill monitoring and intraoperative feedback systems.

Additionally, we observed a recurring pattern, where the kinematic model exhibits sharp upward spikes while the vision model simultaneously shows prediction drops, especially during segments labeled as the pulling gesture (G6). This divergence can be attributed to the different sensitivities of each modality. From the kinematic perspective, the pulling motion is relatively smooth and may be interpreted as skillful, leading to higher predicted scores. However, the vision model captures more context: as the needle is pulled through tissue, the skin is often visibly displaced or stretched, sometimes excessively as shown in Figure \ref{fig: pulling sample}, which introduces a perceived risk of laceration. The vision model appears to correctly associate this with a decline in performance quality, hence the drop in its prediction. This highlights the value of visual input in capturing subtle but clinically relevant indicators of technique that purely kinematic-based data might miss.

\subsection{LOSI analysis}

While LOSO and LOUO are used in the literature for intra-task generalization \cite{hendricks2024exploring}, we proposed the LOSI cross-validation to analyze the learning of skill levels within the JIGSAWS suturing dataset. This approach allows the investigation of skill structures within the suturing and prioritizes data from datasets during training for optimal model training. This analysis approach relies on the idea that datasets may be imbalanced, potentially reflecting the difficulty of recruiting expert surgeons or an over-representation of novice participants and that, as robotic surgery becomes more widespread, larger datasets will need prioritization of the most useful data.

\begin{figure}[t] 
\centering 
\includegraphics[height=5.6cm, trim={3.8cm 0.0cm 4.2cm 0.8cm },clip]
{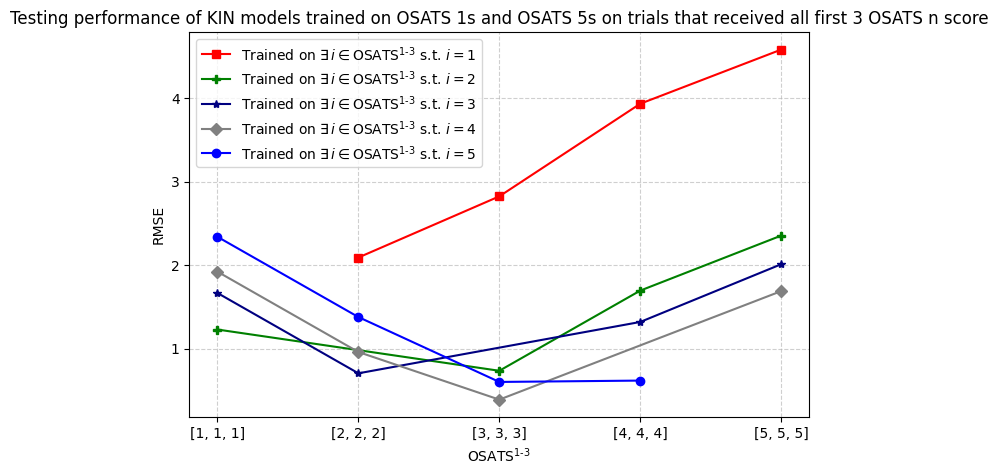}
\caption{Results from LOSI cross validation using kinematic-only model LSTM-K with testing on OSATS$^{1\text{-}3}=[n,n,n]$.}
\label{fig: LOSI kin}
\end{figure}

\begin{figure}[t] 
\centering 
\includegraphics[height=5.6cm, trim={3.8cm 0.0cm 4.2cm 0.8cm },clip]
{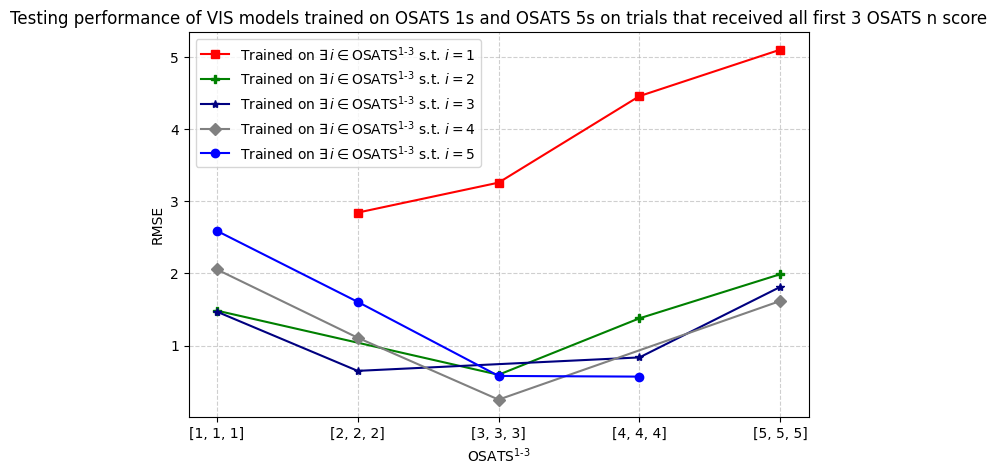}
\caption{Results from LOSI cross validation using vision-only model CNN-LSTM-V with testing on OSATS$^{1\text{-}3}=[n,n,n]$.
} 
\label{fig: LOSI vis}
\end{figure}

We apply LOSI cross-validation, where the model is trained on trials that share at least one common score among the first three OSATS dimensions (i.e. OSATS$^{1\text{-}3}$). Essentially, we train a model on trials that contain a given score in any of the first three OSATS parameters, and evaluate it on the remaining trials, while testing on trials that have all same values in OSATS$^{1\text{-}3}$. We plot the $RMSE$ performance of the five resulting models across these conditions to assess generalization across skill levels. Figure \ref{fig: LOSI kin} and Figure \ref{fig: LOSI vis} present the results obtained using the kinematic and vision-based models, respectively.
The motivation for training on trials that contain a given score in any of OSATS$^{1\text{-}3}$, rather than requiring all three to match, is that scores typically cluster closely for a given participant. This setup helps to maximize training of participants' trials and account for potential reviewer bias, as evaluators often assign similar scores across categories.

The trends that can be extracted from the results indicate that models trained on trials with a score of 1 in OSATS$^{1\text{-}3}$ perform the worst, struggling to generalize to participants with higher skill levels.
Interestingly, this is not true for models trained with a score of 5 in the first three OSATS dimensions, which instead generalize effectively on participants with similar skill level, with the performance declining with participants of lower skill level.
As the result is true for both models trained on kinematic and video data, this suggests that the skill level is detectable in both information channels and that data from higher skill level users have greater importance for skill level detection.
This is likely due to the fact that robotic suturing requires a structured and consistent gesture pattern that is encoded in participants with a high skill level.
This likely reflects the lack of consistent patterns in low-skill performances. 
However, as even low-skilled participants (except those with OSATS equal to 1) showed a similar trend, this indicated that even moderate skill levels exhibit structured gestures that models can learn from.
An additional general trend shows that models trained on adjacent skill levels tend to better predict intermediate trials, with those trained on higher scores often outperforming those trained on lower ones, with the model trained on OSATS$^1$ always scoring the worst. For instance, in the kinematics result (Figure \ref{fig: LOSI kin}), for trials with OSATS$^{1\text{-}3}$ scores of 3-3-3, the model trained on score 4 yields better RMSE than the one trained on score 2, followed by models trained on 5 and then 1, resulting in a performance order of 4 $>$ 2 $>$ 5 $>$ 1.


\section{Conclusion and Future Work}

This paper explored real-time surgical skill assessment by developing unimodal and multimodal models that integrate kinematic and video data. To understand how performance can be enhanced for OSATS predictions in a real-time setup, we investigated: (1) the use of deep learning architectures with different data channels, (2) real-time prediction correlation with surgeons' gestures, and (3) cross-participant generalization across different skill levels.

The results provide insight into the effectiveness of multimodal models for assessing surgical skill using both kinematic and video data. Quantitatively, the mean Spearman correlation coefficients demonstrate that fusion models are more consistent than unimodal architectures, suggesting that combining visual and motion cues provides a more robust representation of surgical performance. Notably, models trained on trials labeled with higher scores in OSATS$^{1\text{-}3}$ generalize better across participants, while those trained on low-skill examples perform poorly, likely due to a lack of structured gesture patterns in novice executions.

Real-time prediction analysis further supports these trends. Visualization of trials' performances shows that the fusion model yields smoother and more accurate temporal predictions, closely aligning with ground truth labels. These predictions are more stable and context-sensitive, particularly during gesture transitions. Interestingly, we observe gesture-specific behaviors: kinematic predictions often spike during pulling gestures, while vision predictions dip, indicating that, while motion data may interpret pulling as smooth and high-skill behavior, visual input can capture risky tissue deformation missed by kinematics alone.

Our LOSI results show that models trained with the lowest score trials (i.e. score 1) perform poorly when predicting higher-skill trials, highlighting the limitations of learning from under-performing demonstrations alone. In contrast, models trained on any higher scores generalize better to neighboring and lower scores. 
This suggests that using data from participants with at least some degree of proficiency is more beneficial than relying solely on those with the lowest skill level, underscoring the importance of including skilled individuals in training datasets.
We also observe that trials with certain scores are best predicted by models trained on adjacent scores, with models trained on higher scores generally outperforming those trained on lower ones. This suggests an asymmetry in generalization: high-skill demonstrations offer richer learning signals for cross-level prediction than low-skill ones.
These findings emphasize the importance of balanced training sets that include mid-to-high scoring trials, especially in automated skill assessment systems. 
It is worth noting that some of these observations might be dataset-specific due to the dataset's size.

There are several avenues that can be addressed in future work. The single-label-per-trial annotation setup limits temporal resolution and introduces potential biases from subjective expert scoring. Future work should involve more granular annotations, for instance, surgeon-provided scores for every 10 seconds window, to better evaluate the temporal validity of real-time predictions. Additionally, the suturing dataset size restricted our analysis primarily to the first three OSATS dimensions. As larger, more diverse datasets become available, future models should aim to predict all six OSATS dimensions and explore differences between annotators to mitigate label bias.



\bibliographystyle{IEEEtran}
\bibliography{bibliography}

\vfill

\end{document}